\documentclass[journal]{IEEEtran}
\usepackage[T1]{fontenc}
\usepackage{cite}
\usepackage{amsmath,amssymb,amsfonts}
\usepackage{tcolorbox}
\tcbuselibrary{breakable}
\usepackage{graphicx}
\usepackage{textcomp}
\usepackage{xcolor}
\usepackage{booktabs} 
\usepackage{multirow} 
\usepackage{algorithm}
\usepackage{algpseudocode}
\usepackage{url}
\usepackage{balance}  
\definecolor{ieeeBlue}{RGB}{0, 114, 188}
\definecolor{ieeeRed}{RGB}{214, 39, 40}
\definecolor{ieeeGray}{RGB}{128, 128, 128}
\definecolor{lightYellow}{RGB}{255, 253, 231}
\definecolor{colorTeal}{RGB}{0, 114, 187}
\definecolor{colorTealLight}{RGB}{208, 232, 242}
\definecolor{colorAmber}{RGB}{243, 156, 18}
\definecolor{colorAmberLight}{RGB}{253, 237, 208}
\definecolor{colorRed}{RGB}{192, 57, 43}
\definecolor{colorGray}{RGB}{127, 140, 141}

\usepackage{tikz}
\usetikzlibrary{shapes.geometric, arrows.meta, positioning, fit, calc, backgrounds, shadows}

\tikzstyle{io} = [trapezium, trapezium left angle=70, trapezium right angle=110, minimum width=1.5cm, minimum height=1cm, text centered, draw=black, fill=blue!5]
\tikzstyle{process} = [rectangle, minimum width=2cm, minimum height=1cm, text centered, draw=black, fill=orange!10, rounded corners]
\tikzstyle{decision} = [diamond, minimum width=2cm, minimum height=1cm, text centered, draw=black, fill=green!10, aspect=2]
\tikzstyle{arrow} = [thick,->,>=stealth]
\tikzstyle{group} = [draw=gray, dashed, inner sep=0.5cm, rounded corners]
\begin{document}

\title{Proactive Rejection  and Grounded Execution: A Dual-Stage Intent Analysis Paradigm for Safe and Efficient AIoT Smart Homes}


\author{Xinxin~Jin, Zhengwei~Ni,~\IEEEmembership{Member,~IEEE,} Zhengguo Sheng,~\IEEEmembership{Senior~Member,~IEEE,}  and Victor~C.~M.~Leung,~\IEEEmembership{Life Fellow,~IEEE}
	\thanks{The work of Zhengwei Ni was supported in part by Zhejiang Provincial Natural Science Foundation of China under Grant No.LMS26F010017 and in part by the Tongxiang General Artificial Intelligence Research Institute Project TAGI2-B-2024-0017.}
	\thanks{Xinxin Jin and Zhengwei Ni are with the School of Information and Electronic Engineering (Sussex Artificial Intelligence Institute), Zhejiang Gongshang University, Hangzhou 310018, China. (e-mail: 24020090056@pop.zjgsu.edu.cn, zhengwei.ni@zjgsu.edu.cn)}
	\thanks{Zhengguo Sheng is with the Sussex Artificial Intelligent Institute, Zhejiang Gongshang University, Hangzhou 310018, China, and also with the Department of Engineering and Design, University of Sussex, BN1 9RH, Brighton, U.K. (e-mail: z.sheng@sussex.ac.uk)}
	\thanks{Victor C. M. Leung is with the Artificial Intelligence Research Institute, Shenzhen MSU-BIT University, Shenzhen 518172, China, also with the College of Computer Science and Software Engineering, Shenzhen University, Shenzhen 518060, China, and also with the Department of Electrical and Computer Engineering, The University of British Columbia, Vancouver, BC V6T 1Z4, Canada (e-mail: vleung@ieee.org).}
	\thanks{Zhengwei Ni is the corresponding author.}

}

\maketitle

\begin{abstract}
As Large Language Models (LLMs) transition from information providers to embodied agents in the Internet of Things (IoT), they face significant challenges regarding reliability and interaction efficiency. Direct execution of LLM-generated commands often leads to entity hallucinations (e.g., trying to control non-existent devices). Meanwhile, existing iterative frameworks (e.g., SAGE) suffer from the Interaction Frequency Dilemma, oscillating between reckless execution and excessive user questioning. To address these issues, we propose a Dual-Stage Intent-Aware (DS-IA) Framework. This framework separates high-level user intent understanding from low-level physical execution. Specifically, Stage 1 serves as a semantic firewall to filter out invalid instructions and resolve vague commands by checking the current state of the home. Stage 2 then employs a deterministic cascade verifier—a strict, step-by-step rule checker that verifies the room, device, and capability in sequence—to ensure the action is actually physically possible before execution. Extensive experiments on the HomeBench and SAGE benchmarks demonstrate that DS-IA achieves an Exact Match (EM) rate of 58.56\% (outperforming baselines by over 28\%) and improves the rejection rate of invalid instructions to 87.04\%. Evaluations on the SAGE benchmark further reveal that DS-IA resolves the Interaction Frequency Dilemma by balancing proactive querying with state-based inference. Specifically, it boosts the Autonomous Success Rate (resolving tasks without unnecessary user intervention) from 42.86\% to 71.43\%, while maintaining high precision in identifying irreducible ambiguities that truly necessitate human clarification. These results underscore the framework's ability to minimize user disturbance through accurate environmental grounding.
\end{abstract}


\section{Introduction}
\label{sec:introduction}

With the exponential expansion of the Internet of Things (IoT) ecosystem and the deep integration of Artificial Intelligence (AI), smart home agents are evolving from rule-based traditional models to AI-driven autonomous modes. In this evolution, Large Language Models (LLMs), with their exceptional natural language understanding, commonsense reasoning{\cite{cot}}, and generalized knowledge, have become the core driver for the next generation of intelligent IoT \cite{ref1}. This technological advancement has propelled a paradigm shift in IoT applications: from the traditional ``Ask-Answer'' process to an ``Ask-Act'' process. In the Ask-Act phase, LLMs are no longer merely information providers but are required to function as Embodied Agents, generating specific action plans based on user instructions and environmental states to drive physical devices \cite{sage}. However, while this shift grants smart homes immense flexibility, it introduces fundamental challenges: due to the inherent stochasticity of LLM-generated content, connecting them directly via APIs to the deterministic physical world creates a natural ``Reliability Gap,'' leading to execution errors and even physical safety risks.

To bridge this gap and achieve reliable grounded control, existing research has primarily explored two interaction reasoning modes, both showing theoretical limitations when handling complex real-world home instructions. The first is the single-turn interaction mode. Although Ren et al. proposed ``Uncertainty Alignment'' to seek human help when confidence is low \cite{ren2023}, our research finds that relying solely on model self-consistency to judge physical ambiguity is infeasible. Lacking environmental state constraints, single-turn generation often exhibits over-confidence, resulting in severe Entity Hallucinations. The second mode is the iterative pattern represented by SAGE (Iterative ReAct) \cite{sage}, which attempts to solve execution problems through a loop of ``Instruction-Plan-Tool Use.'' However, SAGE faces a severe ``Interaction Frequency Dilemma'' in deployment: the system struggles to balance ``silent execution'' and ``active questioning.'' Conservative strategies tend to frequently question the user, disrupting the convenience of smart homes, while aggressive strategies may execute recklessly to minimize interaction.

\begin{figure*}[t]
\centering
\begin{tikzpicture}[
    node distance=0.8cm and 0.8cm, 
    font=\sffamily\footnotesize, 
    >=Latex,
    userinput/.style={
        trapezium, trapezium left angle=75, trapezium right angle=105,
        draw=ieeeBlue, fill=ieeeBlue!10, thick,
        align=center, text width=4cm, minimum height=0.8cm,
        inner xsep=2pt
    },
    process/.style={
        rectangle, draw=black!70, fill=white, thick,
        align=center, text width=4.5cm, minimum height=0.8cm, rounded corners=2pt
    },
    interaction/.style={
        rectangle, draw=orange!80, fill=orange!5, thick, dashed,
        align=center, text width=4.5cm, minimum height=0.7cm
    },
    envstate/.style={
        rectangle, draw=ieeeGray, fill=ieeeGray!10, thick, dashed,
        align=left, text width=3cm, font=\scriptsize
    },
    failnode/.style={
        rectangle, draw=ieeeRed, fill=ieeeRed!10, line width=1.2pt,
        text=ieeeRed, font=\sffamily\bfseries,
        align=center, text width=4.5cm, minimum height=0.8cm, rounded corners=2pt
    },
    calloutbox/.style={
        rectangle, draw=orange!80, fill=lightYellow, thick,
        align=center, text width=4.5cm, font=\scriptsize\itshape, rounded corners, inner sep=4pt
    },
    stdarrow/.style={->, thick, draw=black!70},
    failarrow/.style={->, line width=1.5pt, draw=ieeeRed}
]

\node[userinput] (A1) at (0,0) {User Input:\\"Turn on reading lamp,\\\textbf{AND} lock front door."};
\node[font=\bfseries\normalsize, above=0.3cm of A1] (titleA) {(a) Task Omission (Myopic)};

\node[process, below=of A1] (A2) {Agent Processing:\\Focus on Sub-task 1};
\node[process, below=of A2] (A3) {Ambiguity Check:\\Found multiple lamps?};
\node[interaction, below=of A3] (A4a) {Tool Call:\\\texttt{ask\_user("Which one?")}};
\node[userinput, below=of A4a, fill=green!10, draw=green!60!black, text width=3.5cm] (A4b) {User Answer:\\"The one in the bedroom."};
\node[process, below=of A4b] (A5) {Execute Action:\\\texttt{turn\_on(bedroom\_lamp)}};
\node[failnode, below=of A5] (A6) {Premature End of Execution\\(Sub-task 2 forgotten!)};
\node[calloutbox, below=0.4cm of A6] (CalloutA) {Gets trapped in tool details;\\Sub-task 2 is completely dropped!};

\draw[stdarrow] (A1) -- (A2);
\draw[stdarrow] (A2) -- (A3);
\draw[stdarrow] (A3) -- node[right, font=\scriptsize]{Yes} (A4a);
\draw[stdarrow] (A4a) -- (A4b);
\draw[stdarrow] (A4b) -- (A5);
\draw[failarrow] (A5) -- (A6);
\draw[->, thick, dashed, orange!80] (A6) -- (CalloutA);

\def\panelBoffset{7.5cm} 

\node[userinput] (B1) at (\panelBoffset, 0) {User Input:\\"Turn on the \textbf{humidifier}\\in the kitchen."};
\node[font=\bfseries\normalsize, above=0.3cm of B1] (titleB) {(b) Forced Hallucination};

\node[process] (B2) at (B1 |- A2) {Agent Processing:\\Attempt to locate\\\texttt{kitchen.humidifier}};
\node[envstate, right=0.3cm of B2, anchor=west, yshift=0.6cm] (B_Env) {\textbf{Environment:}\\Kitchen: [Oven, Fridge]\\(No humidifier!)};
\draw[dashed, thick, draw=ieeeGray!80, ->] (B_Env.west) -- ++(-0.2,0) |- (B2.east);

\node[process] (B3) at (B1 |- A3) {Entity Check:\\Target not found?};
\node[interaction] (B4) at (B1 |- A4a) {Blind Exploration Tool:\\\texttt{search\_device("humidifier")}};
\node[process] (B5) at (B1 |- A5) {Incorrect Mapping:\\Found \texttt{living\_room.humidifier}};
\node[failnode] (B6) at (B1 |- A6) {Forced Execution\\(on wrong device):\\\texttt{turn\_on(living\_room...)}};
\node[calloutbox] (CalloutB) at (B1 |- CalloutA) {Fails to reject invalid entity;\\Forcibly maps to a wrong device!};

\draw[stdarrow] (B1) -- (B2);
\draw[stdarrow] (B2) -- (B3);
\draw[stdarrow] (B3) -- node[right, font=\scriptsize]{Yes, try fixing} (B4);
\draw[stdarrow] (B4) -- (B5); 
\draw[failarrow] (B5) -- (B6);
\draw[->, thick, dashed, orange!80] (B6) -- (CalloutB);

\begin{pgfonlayer}{background}
    \node[fit=(titleA)(A1)(CalloutA)(A6), draw=black!20, dashed, rounded corners, inner sep=12pt, fill=black!2] {};
    \node[fit=(titleB)(B1)(B_Env)(CalloutB)(B6), draw=black!20, dashed, rounded corners, inner sep=12pt, fill=black!2] {};
\end{pgfonlayer}
\end{tikzpicture}
\caption{Failure mechanisms in existing reactive frameworks. (a) Task Omission: The agent loses global context when trapped in multi-turn interactions. (b) Forced Hallucination: The agent forcibly maps an unexecutable command to an incorrect entity due to the lack of a proactive rejection mechanism.}
\label{fig:reactive_failures}
\end{figure*}
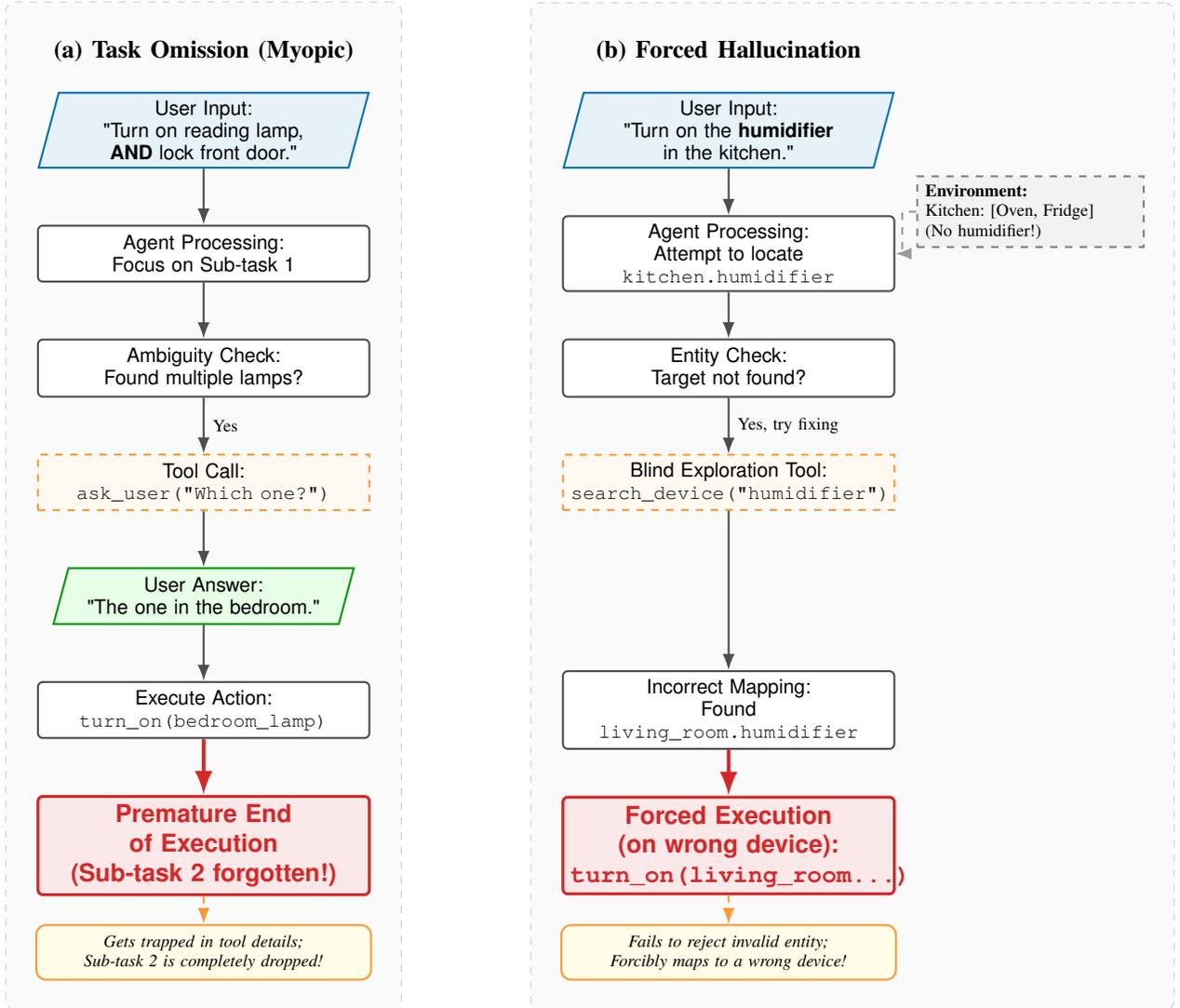

The root cause of this dilemma lies in the lack of deep cognition of instruction ambiguity and global planning in existing iterative frameworks. Ivanova et al., in the AmbiK dataset, classified home instruction ambiguity into ``preference-based,'' ``common-sense,'' and ``safety-critical'' types \cite{ambik}. Frameworks like SAGE, adopting a ``think-while-acting'' reactive strategy, lack mechanisms to identify these ambiguity types before execution. This leads to ``myopic'' behavior, where the model gets trapped in tool invocation details. This blind exploration causes models to fail in distinguishing between preference ambiguities (requiring confirmation) and invalid entity errors (requiring rejection) when facing mixed instructions, leading to two typical failures, as visually demonstrated in Fig. \ref{fig:reactive_failures}: Task Omission (forgetting subsequent instructions) and Forced Hallucination (forcibly mapping to incorrect devices).

Addressing the contradiction between ``unconstrained single-turn generation'' and ``myopic iterative interaction,'' we draw inspiration from Zhang et al.'s use of intent analysis for jailbreak defense \cite{zhang2024} and propose establishing a ``Global Decision Context.'' We construct a Dual-Stage Intent-Aware (DS-IA) Framework designed to combine global perspective with rigorous execution. Unlike SAGE's passive ``error-then-fix'' mechanism, our framework adopts a ``Proactive Analysis'' paradigm. Specifically, Stage 1 is Global Intent Analysis, acting as a ``semantic router'' to assess validity and strip core intents; Stage 2 is Grounding Validation, introducing a ``Room-Device-Capability'' cascade check and a ``Generate-and-Filter'' strategy for mixed intents, ensuring every output action is anchored to a real physical entity.

Our main contributions are summarized as follows:
\begin{itemize}
    \item \textbf{Proposing an ``Analyze-then-Act'' Proactive Paradigm:} We reveal the mechanism behind the ``Interaction Frequency Dilemma'' in existing iterative frameworks and propose a mechanism that decouples macro-intent analysis from micro-action execution.
    \item \textbf{Designing a Dual-Stage Framework with Cascade Verification:} We construct a system architecture featuring ``Pre-execution Intent Routing'' and ``Hierarchical Entity Verification.'' Specifically for mixed intents, we propose a ``Generate-and-Filter'' strategy to eliminate task omission and forced hallucinations.
    \item \textbf{Validation on HomeBench \cite{homebench} and SAGE Benchmarks:} Experiments on the challenging HomeBench dataset show our method significantly reduces hallucination rates. Furthermore, evaluations on the SAGE benchmark prove that our method minimizes unnecessary human interactions while maintaining high task success rates.
\end{itemize}
To facilitate future research and foster open collaboration, the complete codebase, environment snapshots, and evaluation scripts used in this study are publicly available at \url{https://github.com/xxscy/DS-IA-smart.git}.

\section{Related Work}
\label{sec:related_work}

This study intersects with three key domains: LLM-based planning for IoT, ambiguity resolution in human-robot interaction, and efficient agent deployment.

\subsection{LLM-based Planning for Smart Homes}
The paradigm of smart home control is shifting from slot-filling to generative planning. Early works like \cite{party2023} explored utilizing LLMs to infer high-level user intents (e.g., "set up for a party") into device states. To bridge the gap between generation and execution, ProgPrompt \cite{progprompt} introduced a programmatic prompt structure that grounds plans into pythonic code, significantly improving executability. Similarly, IoT-LLM \cite{iotllm} demonstrated that augmenting LLMs with IoT sensor data can enhance physical reasoning capabilities.

However, these methods typically operate in an open-loop manner or rely on heavy prompt engineering. SAGE \cite{sage} advanced this by introducing an iterative ReAct loop. Yet, as noted in recent studies on IoT-LM \cite{iotlm}, purely reactive agents struggle with multi-sensory alignment and often incur high latency. Our DS-IA framework builds upon these foundations but introduces a dual-stage verification mechanism to enforce physical grounding constraints explicitly.

\subsection{Ambiguity Resolution and Human Interaction}
Ambiguity is a pervasive challenge in domestic environments. The AmbiK dataset \cite{ambik} categorizes ambiguity into preference, common-sense, and safety types. Existing approaches to resolve these ambiguities largely rely on human intervention or multimodal cues. For instance, Abugurain et al. \cite{integrating2024} proposed a framework that uses GPT-4 to generate clarifying questions for users, while Calò et al. \cite{enhancing2024} utilized visual cues to disambiguate commands.

While effective, these ``Ask-First'' strategies contribute to the ``Interaction Frequency Dilemma.'' Frequent questioning can degrade user experience. Our work proposes a \textit{State-Aware Disambiguation} strategy, which aligns with the goal of reducing human cognitive load by inferring intent from the current environmental snapshot, calling for help only when necessary.

\subsection{Security, Privacy, and Efficiency}
Deploying LLMs in private homes raises concerns regarding hallucinations and data privacy. ChatIoT \cite{chatiot} highlights the security risks in IoT management and proposes Retrieval-Augmented Generation (RAG) to mitigate threats. Furthermore, to address the computational overhead of large models, Huang et al. \cite{privacy2025} explored Tailored Small Language Models (SLMs) for privacy-preserving on-device processing. These works inspire our future direction of distilling the Intent Analysis module into edge-friendly SLMs to balance performance and privacy.
\begin{table*}[htbp]
\centering
\caption{Comparison of Smart Home Control Frameworks. DS-IA achieves the best balance between physical safety and interaction efficiency.}
\label{tab:related_work_comparison}
\resizebox{\textwidth}{!}{
\begin{tabular}{l|c|c|c|c|c}
\toprule
\textbf{Framework Type} & \textbf{Representative Methods} & \textbf{Reasoning Paradigm} & \textbf{Ambiguity Strategy} & \textbf{Physical Safety} & \textbf{Main Limitation} \\
\midrule
Slot Filling & BERT-based & Supervised / Template & None & High (Template-bound) & Poor generalization to unseen commands \\
\midrule
Direct Generation & GPT-3.5/4 & End-to-End & Model Probability & Low (Prone to Hallucination) & Severe entity hallucinations; no feedback \\
\midrule
Reactive Agents & SAGE \cite{sage}, ReAct \cite{react} & Iterative (Think-Act Loop) & Passive (On Error) & Medium & \textbf{Interaction Frequency Dilemma}; Disturbing \\
\midrule
Uncertainty Planning & Ren et al. \cite{ren2023} & Probabilistic Alignment & Confidence Threshold & Medium & Linguistic confidence $\neq$ Physical truth \\
\midrule
\textbf{DS-IA (Ours)} & \textbf{This Work} & \textbf{Analyze-then-Act (Dual-Stage)} & \textbf{Active State-Awareness} & \textbf{Very High (Cascade Check)} & \textbf{Dependent on real-time snapshot freshness} \\
\bottomrule
\end{tabular}
}
\end{table*}
\section{Methodology}
\label{sec:methodology}

\begin{figure*}[htbp]
\centering
\includegraphics[width=0.95\textwidth]{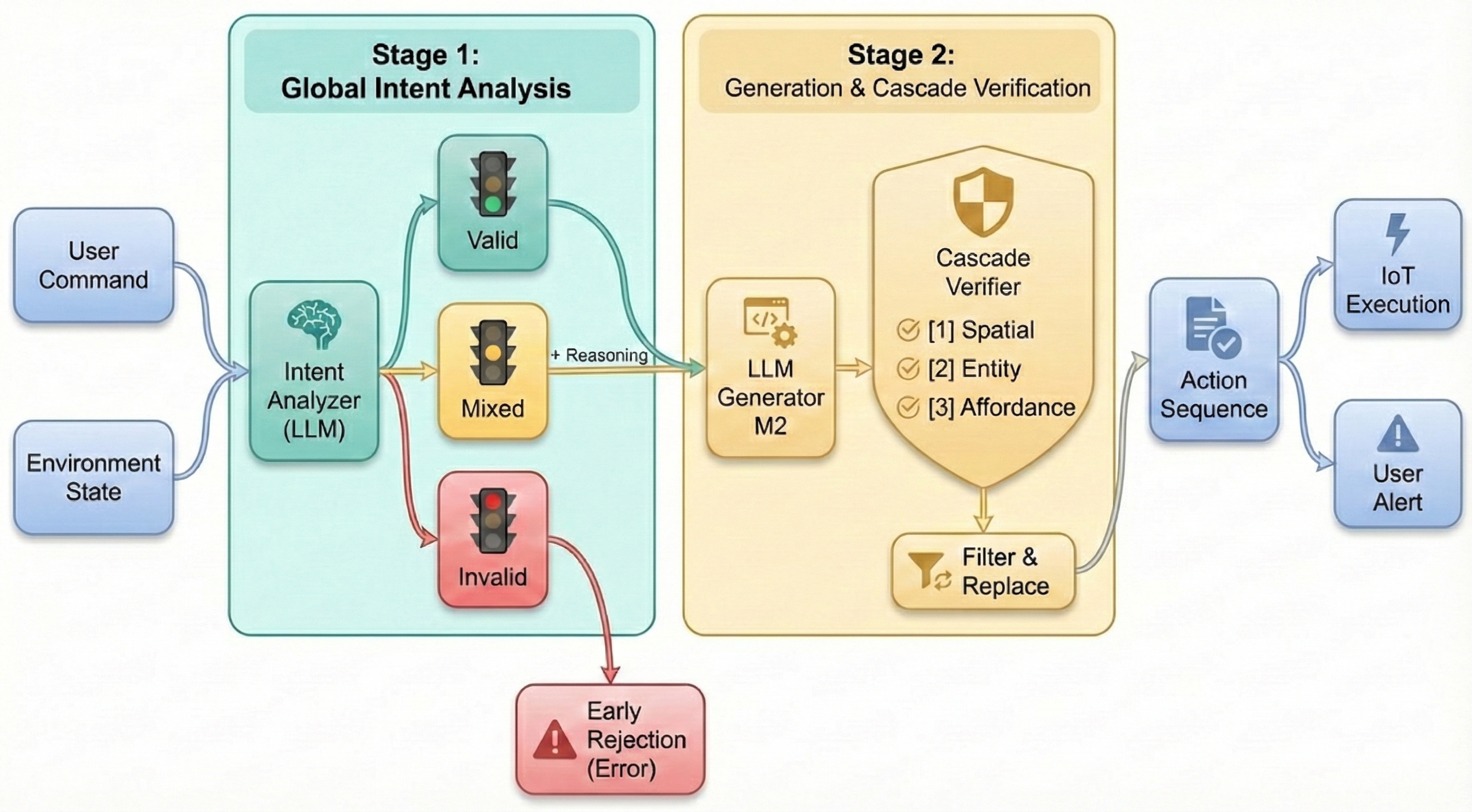} 
\caption{Workflow of the Dual-Stage Intent-Aware (DS-IA) Framework. This image shows the modular architecture: Stage 1 proactively analyzes intent, and Stage 2 utilizes a cascade verifier shield to ensure physical feasibility.}
\label{fig:framework}
\end{figure*}

\subsection{Problem Formulation}
We model the smart home instruction execution task as a Physically Constrained Sequence Generation Problem.

\textbf{Definition 1 (Environment State $S_t$):} We formalize the home environment state at time $t$ as a hierarchical knowledge graph $S_t = \{R, D, C\}$.
\begin{itemize}
    \item $R = \{r_1, ..., r_n\}$ represents the set of physical rooms.
    \item $D_t = \{(d_i, r_{d_i}) \mid d_i \in \mathcal{D}_{global}, r_{d_i} \in R\}$ represents the set of available devices bound to specific rooms at time $t$, where $\mathcal{D}_{global}$ denotes the universal vocabulary of all supported device types in the system.
    \item $C_t(d_i) = \{f_{i,1}, ..., f_{i,k}\}$ represents the set of callable capabilities (methods) currently available for device $d_i$ at time $t$.
\end{itemize}

\textbf{Definition 2 (Instruction and Action):} Given a user instruction $U$, the system generates an action sequence $A = \{a_1, a_2, ..., a_m\}$. Each atomic action $a_k$ is represented as a structured tuple:
\begin{equation}
a_k = \langle r_k, d_k, f_k, p_k \rangle
\end{equation}
where $r_k$, $d_k$, $f_k$, and $p_k$ denote the target room, target device identifier, target capability (function), and function parameters, respectively.

Traditional end-to-end models attempt to maximize the conditional probability $P(A|U; \theta)$, where $\theta$ represents the \textbf{frozen parameters} of a pre-trained LLM. However, in open-vocabulary settings, this unconstrained probabilistic generation often leads to $a_k$ pointing to non-existent entities. Thus, we introduce a physical grounding constraint function $\Phi(A, S_t)$ and modify the \textbf{inference-time} decoding objective to:
\begin{equation}
A^* = \operatorname*{argmax}_{A} \left( P(A|U, S_t; \theta) \cdot \mathbb{I}[\Phi(A, S_t) = 1] \right)
\end{equation}
where $\mathbb{I}[\cdot]$ is the indicator function that outputs 1 if the generated sequence is entirely feasible within the current environment, and 0 otherwise.

\subsection{Stage 1: Global Intent Analysis and Routing}
The Intent Analysis (IA) module acts as a semantic firewall. It utilizes the LLM's semantic understanding to parse the instruction based strictly on the current environment snapshot $S_t$. The output of this module is formalized as an analysis object $\mathcal{O}_{IA} = \langle \text{Type}, \text{Reasoning} \rangle$, where $\text{Type}$ categorizes the instruction into one of three classes, and $\text{Reasoning}$ contains the diagnostic context. The three classes are:
\begin{itemize}
    \item $C_{valid}$ (Valid): All mentioned entities can be found in $S_t$ with high confidence.
    \item $C_{invalid}$ (Invalid): The instruction explicitly points to non-existent devices. The system triggers Early Rejection.
    \item $C_{mixed}$ (Mixed): The instruction contains multiple sub-tasks with a mix of valid and invalid entities.
\end{itemize}

\subsection{Stage 2: Hierarchical Grounding Verification}
For instructions passed by IA ($C_{valid}$ and $C_{mixed}$), we use In-Context Learning to generate a raw candidate action sequence $A_{raw}$. To eradicate hallucinations, we design a Three-Level Cascade Verifier to inspect each atomic action $a_k = \langle r_k, d_k, f_k, p_k \rangle \in A_{raw}$:

\textbf{Level 1: Spatial Topology Verification ($V_R$)}
Checks if the target room exists in the house:
\begin{equation}
V_R(a_k) = 1 \iff r_k \in R
\end{equation}

\textbf{Level 2: Entity Alignment Verification ($V_D$)}
Checks if the target device actually exists in the specified room at time $t$:
\begin{equation}
V_D(a_k) = 1 \iff (d_k, r_k) \in D_t
\end{equation}

\textbf{Level 3: Affordance Verification ($V_C$)}
Checks if the device supports the requested capability at time $t$:
\begin{equation}
V_C(a_k) = 1 \iff f_k \in C_t(d_k)
\end{equation}

The atomic constraint for a single action is $\Phi(a_k, S_t) = V_R(a_k) \land V_D(a_k) \land V_C(a_k)$. To strictly align with standard embodied evaluation protocols (such as the HomeBench dataset), if an action fails any of these checks ($\Phi(a_k, S_t) = 0$), the verifier explicitly outputs a standardized error token ($\epsilon_{err}$) to flag the physical violation. Consequently, the sequence-level constraint from Eq. 2 is enforced by applying this atomic filtering across all generated steps.

\subsection{Mixed Intent Resolution Strategy}
For $C_{mixed}$ instructions, where valid and invalid sub-tasks coexist (e.g., "turn on the TV and the non-existent heater"), traditional reactive models typically suffer from either ``all-or-nothing'' execution failures or forced hallucinations. To overcome this, we employ a ``Generate-and-Filter'' strategy (Algorithm \ref{alg:mixed_intent}) designed to maximize task recall while strictly preventing unexecutable actions.

The operational process consists of two main phases:
\begin{enumerate}
    \item \textbf{Reasoning-Guided Generation:} Instead of generating actions blindly, the Stage 2 LLM generator drafts a raw candidate sequence $A_{raw}$ conditioned on the diagnostic context ($\mathcal{O}_{IA}.\text{Reasoning}$) explicitly passed from Stage 1. This prior knowledge serves as a macro-level roadmap, pre-warning the generator about missing entities and reducing the search space.
    \item \textbf{Atomic Filtering and Sequential Alignment:} Once $A_{raw}$ is generated, the Three-Level Cascade Verifier independently evaluates each atomic action $a_k$. As required by standard benchmark protocols, unexecutable actions must be flagged with the error token ($\epsilon_{err}$). In the context of complex mixed intents, our framework perfectly masters this requirement: instead of suffering from cascading execution failures or blindly dropping the failed task, the verifier precisely substitutes the localized failure with $\epsilon_{err}$. This precise substitution is structurally crucial, as it preserves the sequential alignment of the original multi-step instruction, completely preventing subsequent valid actions from being temporally displaced or entirely omitted (Task Omission). 
\end{enumerate}

Finally, the system aggregates the execution results to formulate precise textual feedback ($Msg$). Rather than returning a generic error, the agent informs the user exactly which sub-tasks succeeded and which were safely bypassed due to physical constraints, thereby ensuring a transparent human-agent interaction.

\subsection{A Running Example of the DS-IA Framework}
\label{subsec:running_example}

To concretely illustrate how the DS-IA framework bridges the reliability gap, consider a complex $C_{mixed}$ instruction and a specific environment snapshot $S_t$:

\begin{itemize}
    \item \textbf{User Instruction ($U$):} \textit{"Turn on the bedroom reading lamp, set the kitchen dehumidifier to 50\%, and lock the front door."}
    \item \textbf{Environment Snapshot ($S_t$):} 
    \begin{itemize}
        \item \texttt{Living Room}: Reading Lamp (ON)
        \item \texttt{Bedroom}: Reading Lamp (OFF)
        \item \texttt{Kitchen}: Oven, Fridge (No Dehumidifier)
        \item \texttt{Entrance}: Smart Lock (UNLOCKED)
    \end{itemize}
\end{itemize}

\textbf{Step 1: Stage 1 Intent Analysis \& Routing.} 
The LLM router analyzes $U$ against $S_t$. For "bedroom reading lamp," it confirms the entity exists and is valid ($C_{valid}$). For "kitchen dehumidifier," the router detects a $C_{invalid}$ intent because the device is missing from the snapshot. For "lock front door," it is $C_{valid}$. The overall instruction is routed as $C_{mixed}$ and passed to Stage 2 with its diagnostic reasoning $\mathcal{O}_{IA}$.

\textbf{Step 2: Stage 2 Generation \& Cascade Verification.} 
The generator drafts a raw action sequence $A_{raw}$. The Cascade Verifier then strictly checks each action:
\begin{enumerate}
    \item \texttt{turn\_on(bedroom\_lamp)}: Passes $V_R \land V_D \land V_C$. (Accepted)
    \item \texttt{set\_humidity(kitchen\_dehum, 50)}: Fails Level 2 Verification ($V_D = 0$). Replaced with error token $\epsilon_{err}$.
    \item \texttt{lock(entrance\_lock)}: Passes all checks. (Accepted)
\end{enumerate}

\textbf{Final Output:} 
Unlike reactive frameworks that might hallucinate a dehumidifier or drop the lock command due to mid-way tool failures, DS-IA safely outputs a filtered sequence $A_{final} = \{\texttt{turn\_on(bedroom\_lamp), lock(entrance\_lock)}\}$ and alerts the user: \textit{"Executed lamps and locks. Failed: Kitchen dehumidifier not found."} This demonstrates zero Task Omission and zero Forced Hallucination.

\subsection{Prompt Engineering Strategy}
The performance of the DS-IA framework relies heavily on structured prompt design. Moving away from standard few-shot prompting, we adopt a ``Constraint-Aware Instruction Prompting'' strategy. We introduce two key mechanisms into the prompts:
\begin{enumerate}
    \item \textbf{State Injection:} We flatten the hierarchical knowledge graph $S_t$ into a compact textual representation.
    \item \textbf{Negative Constraints:} We explicitly instruct the model on \textit{what not to do} (e.g., "Output \texttt{error\_input} when operating non-existent attributes and devices").
\end{enumerate}

Due to space limitations, the exact and complete prompt templates utilized for both Stage 1 (Global Intent Analysis) and Stage 2 (Code Generation) are detailed in Appendix .

\begin{algorithm}[h]
\caption{Mixed Intent Resolution via Generate-and-Filter}
\label{alg:mixed_intent}
\begin{algorithmic}[1]
\Require User Instruction $U$, Environment State $S_t$
\Ensure Executable Sequence $A_{final}$, Feedback $Msg$

\State \textbf{// Stage 1: Intent Routing}
\State $\mathcal{O}_{IA} \leftarrow \text{IntentAnalysis}(U, S_t)$
\If{$\mathcal{O}_{IA}.\text{Type} == C_{invalid}$}
    \State \Return $\emptyset$, "Operation rejected: No valid device."
\EndIf

\State \textbf{// Stage 2: Generation \& Filtering}
\State $A_{raw} \leftarrow \text{LLM\_Generate}(U, S_t, \mathcal{O}_{IA}.\text{Reasoning})$
\State $A_{final} \leftarrow \text{List}()$, $ErrorSet \leftarrow \text{Set}()$

\For{\textbf{each} atomic action $a_k = \langle r_k, d_k, f_k, p_k \rangle$ \textbf{in} $A_{raw}$}
    \State $v_1 \leftarrow V_R(a_k)$
    \State $v_2 \leftarrow V_D(a_k)$
    \State $v_3 \leftarrow V_C(a_k)$

    \If{$v_1 \land v_2 \land v_3$ \textbf{is True}}
        \State $A_{final}.\text{append}(a_k)$
    \Else
        \State $A_{final}.\text{append}(\epsilon_{err})$ \Comment{Error Token}
        \State $ErrorSet.\text{add}(d_k)$
    \EndIf
\EndFor

\State \textbf{// Construct Feedback}
\If{$ErrorSet$ is not empty}
    \State $Msg \leftarrow$ "Executed valid actions. Failed: " + $ErrorSet$
\Else
    \State $Msg \leftarrow$ "Success."
\EndIf

\State \Return $A_{final}, Msg$
\end{algorithmic}
\end{algorithm}

\section{Experimental Setup}
\label{sec:setup}

We evaluate the DS-IA framework on two complementary benchmarks: HomeBench for robustness and SAGE Benchmark for interaction efficiency.

\subsection{Datasets}

\subsubsection{HomeBench (Robustness)}
HomeBench contains 100 simulated home environments and 2,500 instructions, evaluating the agent's ability to ground instructions in physical constraints. The instructions are categorized into three types: Valid (standard executable commands), Invalid (targeting non-existent entities, comprising 38.6\% of the data), and Mixed (containing both valid and invalid sub-tasks).

\vspace{1mm}
\noindent\textbf{Input/Output Instance (Mixed Task):}
\begin{itemize}
    \item Input Context ($S_t$): \texttt{Kitchen: [Light (OFF)]}
    \item Input Instruction ($U$): \textit{"Turn on the kitchen light and the oven."}
    \item Expected Output ($A$): \texttt{\{kitchen.light.turn\_on(), $\epsilon_{err}$\}} \\
    \textit{(Explanation: The model must execute the valid light command while replacing the non-existent oven command with an error token.)}
\end{itemize}

\subsubsection{SAGE Benchmark (Interaction Efficiency)}
This benchmark contains 50 complex tasks focusing on Human-Robot Interaction (HRI), covering scenarios like Device Resolution, Personalization, and Intent Disambiguation. It evaluates whether the agent can appropriately balance autonomous execution and proactive questioning.

\vspace{1mm}
\noindent\textbf{Input/Output Instance (Device Resolution):}
\begin{itemize}
    \item Input Context ($S_t$): \texttt{Bedroom: [Lamp\_A (ON), Lamp\_B (OFF)]}
    \item Input Instruction ($U$): \textit{"Turn on the lamp."}
    \item Expected Output (Autonomous): \texttt{\{bedroom.lamp\_B.turn\_on()\}} \\
    \textit{(Explanation: The agent should silently resolve the ambiguity based on the device states without disturbing the user. If both lamps were OFF, the expected output would shift to an interactive tool call like \texttt{ask\_user("Which lamp?")}.)}
\end{itemize}

\subsection{Baselines and Models}
We compare against:
\begin{enumerate}
    \item \textbf{Standard Few-Shot:} Uses 4-shot prompting with full environment context, representing native LLM capabilities.
    \item \textbf{SAGE (Iterative ReAct):} The state-of-the-art industrial framework using tool calls (\texttt{query\_devices}, \texttt{human\_interaction}) \cite{sage}.
\end{enumerate}

To ensure fair comparison and verify generalization:
\begin{itemize}
    \item For HomeBench: All methods use Qwen-2.5-7B-Instruct on an NVIDIA RTX 5090 (BFloat16, SDPA enabled). DS-IA uses Greedy Decoding ($T=0$).
    \item For SAGE Benchmark: We unify both the baseline and our framework to utilize GPT-4o-mini. This choice is made to ensure strict alignment with the original SAGE paper's experimental settings \cite{sage} for fair reproduction and evaluation.
\end{itemize}

\subsection{Evaluation Metrics}
To comprehensively assess both the physical safety and the interaction efficiency of the smart home agents, we employ distinct metric suites tailored to our two benchmarks:

\textbf{1. Metrics for HomeBench (Physical Grounding \& Safety):}
\begin{itemize}
    \item \textbf{Exact Match (EM):} The primary metric for execution accuracy and safety. For valid instructions (Valid Single/Valid Multi/Mixed Multi), it evaluates whether the generated action sequence perfectly matches the oracle ground truth without any hallucinations. For explicitly invalid instructions (IS/IM), a perfect match dictates that the system successfully generated the designated \texttt{error\_input} token, which represents the system's Rejection Rate.
    \item \textbf{F1-Score:} Provides a granular evaluation of partial correctness by measuring the precision and recall of the predicted rooms, devices, and actions/attributes within the generated code.
\end{itemize}

\textbf{2. Metrics for SAGE Benchmark (Interaction Efficiency):}
\begin{itemize}
    \item \textbf{Task Success Rate (Succ. Rate):} A global evaluation of whether a multi-turn task was ultimately resolved. Unlike standard EM, Succ. Rate is achieved through two distinct, context-dependent interaction dimensions:
    \item \textbf{Autonomous Succ. Rate:} Evaluates performance on clear tasks. It measures the agent's ability to successfully deduce and execute the required actions \textit{without} unnecessarily querying the user (penalizing over-cautiousness and redundant API calls).
    \item \textbf{Clarification Succ. Rate:} Evaluates performance on ambiguous tasks. It measures the agent's ability to correctly halt physical execution and proactively invoke human-centric tools (e.g., \texttt{ask\_user}) to acquire missing information (penalizing forced hallucinations).
\end{itemize}

\section{Results and Analysis}
\label{sec:results}

\subsection{Robustness on HomeBench}

To systematically evaluate the framework's robustness against entity hallucinations, we categorize the HomeBench instructions into five distinct types based on their validity and complexity. Table \ref{tab:homebench_taxonomy} visualizes these categories with concrete examples and maps them to their corresponding evaluation metrics. 

\begin{table}[htbp]
\centering
\caption{Taxonomy of HomeBench Tasks and Metric Mapping. (Assuming $S_t$ does NOT contain an Oven)}
\label{tab:homebench_taxonomy}
\resizebox{\columnwidth}{!}{
\begin{tabular}{l|p{3.5cm}|l}
\toprule
\textbf{Task Category} & \textbf{Instruction Example} & \textbf{Targeted Metric \& Capability} \\
\midrule
\textbf{VS} (Valid Single) & "Turn on the lamp." & \multirow{2}{*}{\shortstack[l]{\textbf{EM \& F1:} Evaluates normal \\ instruction-following recall.}} \\
\textbf{VM} (Valid Multi) & "Turn on lamp, lock door." & \\
\midrule
\textbf{IS} (Invalid Single) & "Turn on the oven." & \multirow{2}{*}{\shortstack[l]{\textbf{Rejection Rate:} Evaluates safety \\ and hallucination suppression.}} \\
\textbf{IM} (Invalid Multi) & "Turn on oven and heater." & \\
\midrule
\textbf{MM} (Mixed Multi) & "Turn on lamp and oven." & \textbf{EM \& F1:} Evaluates filter ability. \\
\bottomrule
\end{tabular}
}
\end{table}

\begin{table*}[t]
\centering
\caption{Performance Comparison on HomeBench (EM: Exact Match, F1: F1-Score). \textbf{Bold} indicates best performance.}
\label{tab:homebench_main}
\resizebox{\textwidth}{!}{
\begin{tabular}{l|cc|cc|cc|cc|cc|cc}
\toprule
\multirow{2}{*}{\textbf{Method}} & \multicolumn{2}{c|}{\textbf{VS (Valid Single)}} & \multicolumn{2}{c|}{\textbf{IS (Invalid Single)}} & \multicolumn{2}{c|}{\textbf{VM (Valid Multi)}} & \multicolumn{2}{c|}{\textbf{IM (Invalid Multi)}} & \multicolumn{2}{c|}{\textbf{MM (Mixed Multi)}} & \multicolumn{2}{c}{\textbf{Overall}} \\
& EM & F1 & EM & F1 & EM & F1 & EM & F1 & EM & F1 & \textbf{EM} & \textbf{F1} \\
\midrule
Baseline & \textbf{66.96\%} & \textbf{66.89\%} & 14.07\% & 14.08\% & 37.55\% & 69.18\% & 0.00\% & 5.82\% & 0.49\% & 33.23\% & 29.98\% & 35.72\% \\
SAGE & 48.63\% & 49.18\% & 29.84\% & 29.84\% & 9.09\% & 47.42\% & 0.00\% & 0.00\% & 1.77\% & 29.80\% & 1.77\% & 30.84\% \\
\textbf{DS-IA (Ours)} & 50.40\% & 50.37\% & \textbf{87.04\%} & \textbf{87.16\%} & \textbf{50.00\%} & \textbf{76.98\%} & \textbf{38.46\%} & \textbf{73.08\%} & \textbf{24.49\%} & \textbf{77.42\%} & \textbf{58.56\%} & \textbf{74.90\%} \\
\bottomrule
\end{tabular}
}
\end{table*}

Understanding this mapping is crucial because success on Valid tasks measures \textit{execution capability}, whereas success on Invalid tasks strictly measures \textit{physical safety} (the ability to trigger Early Rejection). Table \ref{tab:homebench_main} presents the comprehensive quantitative results across these dimensions. DS-IA achieves a remarkable overall success rate (EM) of 58.56\% and an F1-score of 74.90\%, significantly outperforming the Baseline (29.98\% EM) and SAGE (1.77\% EM).

\subsubsection{Analysis of Baseline Performance Anomalies}
It is worth noting that the SAGE framework exhibits an unexpectedly low EM rate (1.77\%) on HomeBench compared to its original paper. 
We attribute this performance degradation to the mismatch between Reactive Architectures and Invalid Instructions. 
SAGE relies on an iterative ``Observe-Reason-Act'' loop. When facing invalid instructions (which constitute 38.6\% of HomeBench), SAGE lacks a mechanism to stop early. 
Instead, it exhaustively invokes search tools (e.g., \texttt{scan\_devices}) until it either hallucinates a semantic match (Forced Grounding) or exhausts the context window. 
This highlights the necessity of our Stage 1 ``Semantic Firewall,'' which filters out these unexecutable queries before they trigger the costly reactive loop.

\subsubsection{Analysis of Hallucination Suppression}
In the IS (Invalid Single) task, which directly tests the capability to reject hallucinations, Baseline only achieves 14.07\%, indicating a strong tendency for ``Forced Grounding.'' In contrast, DS-IA achieves 87.04\%. This quantum leap proves that the Stage 1 intent routing mechanism effectively serves as a semantic firewall.

\subsubsection{Analysis of Mixed Intents}
For MM (Mixed Multi) tasks, DS-IA achieves an F1 score of 77.42\% (Success Rate: 24.49\%), compared to Baseline's F1 of 33.23\%. This validates the ``Generate-and-Filter'' strategy, which allows the agent to execute valid sub-tasks while safely discarding hallucinations, avoiding the ``all-or-nothing'' failure mode common in single-stage models.

\subsubsection{Improvement on Valid Instructions}
Notably, compared to previous iterations, our framework maintains a competitive performance on VS (Valid Single) tasks (50.40\%), narrowing the gap with the Baseline while providing significantly higher safety guarantees. For complex VM (Valid Multi) tasks, DS-IA outperforms the Baseline with a 50.00\% success rate, demonstrating superior capability in handling long-horizon planning.

\subsection{Case Study: Suppression of Forced Hallucination}
\label{subsec:case_study}

To intuitively demonstrate the safety advantage of DS-IA over SAGE on the HomeBench dataset, we analyze a representative failure case involving invalid entities (Case ID: \texttt{home45\_one\_1306}).

\textbf{Scenario Description:}
\begin{itemize}
    \item \textbf{User Instruction:} \textit{"Set the intensity of the dehumidifiers to 0 in the store room."}
    \item \textbf{Environment Context:} The simulated home does not contain a ``store room'' or does not have a dehumidifier in that location. The user's request targets a non-existent entity.
    \item \textbf{Ground Truth:} The system should report an error and execute nothing (\texttt{error\_input}).
\end{itemize}

\textbf{Comparison of Framework Behaviors:}
\begin{enumerate}
    \item \textbf{SAGE (Baseline) Failure:}
    As shown in the SAGE execution logs, the baseline model output: 
    \begin{center}
    \texttt{study\_room.dehumidifiers.set\_intensity(0)}
    \end{center}
    \textbf{Analysis:} Lacking an explicit rejection mechanism, SAGE exhibited ``Forced Grounding.'' It semantically mapped the non-existent ``store room'' to the nearest valid entity ``study room'' and executed the command. This illustrates the "Reliability Gap" where the model prioritizes execution over faithfulness, leading to unintended physical operations.

    \item \textbf{DS-IA (Ours) Success:}
    Our framework output:
    \begin{center}
    \texttt{\{error\_input\}}
    \end{center}
    \textbf{Analysis:} 
    \begin{itemize}
        \item \textbf{Stage 1 (Intent Analysis):} The Intent Router scanned the environment snapshot $S_t$. It detected that the target entity \texttt{(store\_room, dehumidifier)} was missing from the graph.
        \item \textbf{Decision:} The instruction was classified as $C_{invalid}$.
        \item \textbf{Action:} The system triggered the Early Rejection protocol, bypassing the generation stage entirely.
    \end{itemize}
\end{enumerate}

This case strongly validates that incorporating a pre-execution verification stage is essential for preventing LLM hallucinations in safety-critical IoT environments.

\begin{figure}[htbp]
\centering
\begin{tikzpicture}[node distance=1.2cm, font=\small]
    \node (user) [draw, rectangle, rounded corners, fill=blue!5, text width=7.5cm, align=center] {\textbf{User:} "Set dehumidifier to 0 in \underline{store room}"};
    
    \node (sage) [below of=user, xshift=-2.2cm, yshift=-0.3cm] {\textbf{SAGE Framework}};
    \node (sage_proc) [draw, rectangle, below of=sage, fill=red!10, text width=3cm, align=center, minimum height=1.5cm] {Search similar...\\ Found: "Study Room"};
    \node (sage_out) [draw, rectangle, below of=sage_proc, fill=red!20, text width=3cm, align=center, yshift=-0.3cm] {\textbf{Wrong Action:}\\ Turn off \textit{Study Room} Dehumidifier};
    
    \node (dsia) [below of=user, xshift=2.2cm, yshift=-0.3cm] {\textbf{DS-IA Framework}};
    \node (dsia_proc) [draw, rectangle, below of=dsia, fill=green!10, text width=3cm, align=center, minimum height=1.5cm] {Stage 1 Check:\\ Store Room $\notin S_t$\\ -> INVALID};
    \node (dsia_out) [draw, rectangle, below of=dsia_proc, fill=green!20, text width=3cm, align=center, yshift=-0.3cm] {\textbf{Correct Action:}\\ Reject \& Report Error};
    
    \draw [->, thick] (user.south) -- ++(0,-0.3) -| (sage.north);
    \draw [->, thick] (user.south) -- ++(0,-0.3) -| (dsia.north);
    \draw [->] (sage) -- (sage_proc);
    \draw [->] (sage_proc) -- (sage_out);
    \draw [->] (dsia) -- (dsia_proc);
    \draw [->] (dsia_proc) -- (dsia_out);
\end{tikzpicture}
\caption{Visual comparison of handling a non-existent device instruction on HomeBench. SAGE hallucinates a target (Forced Grounding), while DS-IA correctly rejects the command (Early Rejection).}
\label{fig:case_study}
\end{figure}
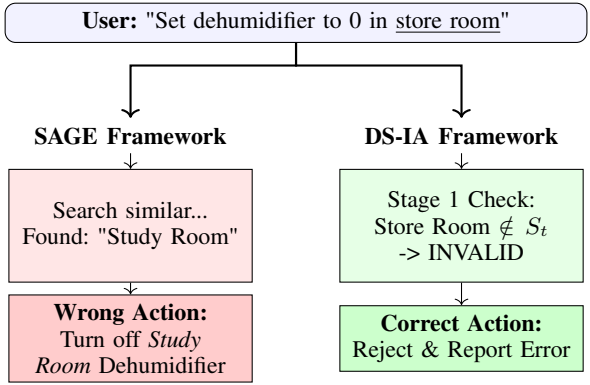

\subsection{Ablation Study: Computational Efficiency and Safety}
\label{subsec:ablation}

To deeply analyze the contribution of the Intent Analysis (IA) module, we conducted an ablation study on a subset of 1,000 tasks, comparing the Full Framework (DS-IA) against a w/o Stage 1 (No IA) baseline. In the baseline, instructions are directly fed to the code generator without prior semantic routing. 

Rather than merely evaluating general execution, this ablation explicitly focuses on the framework's physical safety capabilities and computational economics. Table \ref{tab:ablation_results} details the safety metrics (Invalid Instruction Rejection Rate) alongside the API call frequency and token consumption for both configurations.

\begin{table}[htbp]
\centering
\caption{Ablation on Safety and Efficiency (1,000 Tasks). DS-IA boosts physical safety (Rejection Rate) while trading cheap input tokens to massively reduce expensive autoregressive code generation cycles.}
\label{tab:ablation_results}
\resizebox{\columnwidth}{!}{
\begin{tabular}{l|c|cc|cc}
\toprule
\multirow{2}{*}{\textbf{Method}} & \textbf{Safety} & \multicolumn{2}{c|}{\textbf{Stage 1 (Intent Analysis)}} & \multicolumn{2}{c}{\textbf{Stage 2 (Code Generation)}} \\
& \textbf{IS EM (Rejection)} & \textbf{Calls} & \textbf{Tokens} & \textbf{Calls} & \textbf{Tokens} \\
\midrule
No IA Baseline & 83.29\% & 0 & 0 & 1,000 & 2,982,070 \\
\textbf{DS-IA (Ours)} & \textbf{88.69\%} & 1,000 & 3,262,848 & \textbf{819} & \textbf{2,554,231} \\
\bottomrule
\end{tabular}
}
\end{table}

\subsubsection{Fortifying Safety through Early Rejection}
As observed in Table \ref{tab:ablation_results}, the IA module serves as a critical semantic firewall. Without Stage 1, the baseline attempts to ground every user instruction blindly, leading to a higher risk of Forced Hallucinations on invalid commands. By isolating existence verification in Stage 1, DS-IA correctly intercepts explicitly invalid intents before they reach the generator. This strict \textit{Early Rejection} mechanism boosts the system's absolute physical safety, increasing the exact match rate on Invalid Single (IS) instructions from 83.29\% to 88.69\%, firmly establishing a ``Do No Harm'' protocol for smart home control.

\subsubsection{Massive Reduction in Generation Overhead}
Beyond safety, the Early Rejection mechanism fundamentally optimizes the computational dynamics of the agent. By bypassing the generation phase for explicitly invalid intents, DS-IA slashes the number of heavy Code Generation calls by 18.1\% (from 1,000 down to 819). Consequently, the framework successfully saves over 427,000 generation tokens in Stage 2.

\subsubsection{The Economics of Prefill vs. Decoding}
While DS-IA introduces ~3.26M tokens in Stage 1, this apparent overhead is highly asymmetrical in practice. In modern LLM architectures, Stage 1 (IA) predominantly involves \textit{prefill (prompt) processing} to read the environment snapshot $S_t$. Prefill is highly parallelizable and computationally cheap. Conversely, Stage 2 (Code Generation) relies heavily on \textit{autoregressive decoding}, which is memory-bandwidth-bound, exceptionally slow, and computationally expensive. 

By trading inexpensive input (IA) tokens for a massive reduction in expensive generation (CodeGen) tokens, DS-IA implements a highly efficient Adaptive Computation strategy. It ensures that the heavy lifting of code generation is reserved strictly for physically valid tasks, entirely eradicating the wasted compute on hallucinated commands.
\subsection{Interaction Efficiency on SAGE Benchmark}

To rigorously evaluate interaction efficiency in complex Human-Robot Interaction (HRI) scenarios, we utilize the SAGE Benchmark (50 tasks). Unlike traditional metrics that only reward physical execution, we define Task Success Rate (Succ. Rate) globally: a task is considered successful if the agent either correctly executes an autonomous action when information is sufficient, or accurately invokes human-centric tools (e.g., \texttt{ask\_user}) to proactively acquire missing information when faced with irreducible ambiguity. Table \ref{tab:sage_interaction} illustrates the comprehensive performance across different dimensions.

\begin{table}[htbp]
\centering
\caption{Task Success Rate (Succ. Rate) Breakdown on SAGE Benchmark. Success is achieved by either safe autonomous execution or correct proactive clarification.}
\label{tab:sage_interaction}
\resizebox{\columnwidth}{!}{
\begin{tabular}{l|cc}
\toprule
\textbf{Task Category} & \textbf{SAGE (Baseline)} & \textbf{DS-IA (Ours)} \\
\midrule
\multicolumn{3}{l}{\textit{\textbf{Global Judgment Metrics}}} \\
Clarification Succ. Rate (Ambiguous) & 75.00\% & \textbf{75.00\%} \\
Autonomous Succ. Rate (Clear) & 42.86\% & \textbf{71.43\%} \\
\midrule
\multicolumn{3}{l}{\textit{\textbf{Detailed Task Succ. Rate Breakdown}}} \\
Intent Resolution & 33.33\% & \textbf{70.83\%} \\
Device Resolution & 46.15\% & \textbf{69.44\%} \\
Personalization   & 27.27\% & \textbf{54.55\%} \\
Persistence       & 25.00\% & \textbf{100.00\%} \\
Command Chaining  & 62.50\% & \textbf{62.50\%} \\
Simple Commands   & 57.14\% & \textbf{85.71\%} \\
\bottomrule
\end{tabular}
}
\end{table}

\subsubsection{Comprehensive Analysis of Interaction Data}

\begin{itemize}
    \item \textbf{Balancing Autonomy and Proactive Querying:} At the macro level, SAGE suffers from severe ``Over-cautiousness'' and frequent questioning. When the human interaction module is enabled, the baseline often falls into a repetitive querying loop, achieving an Autonomous Succ. Rate of only 42.86\%. This means it excessively disturbs the user with redundant questions in over half of the tasks that could have been solved independently. By decoupling intent routing from physical generation, DS-IA boosts this autonomy to 71.43\%. Concurrently, on ambiguous tasks strictly requiring human intervention, DS-IA perfectly maintains the Clarification Succ. Rate at 75.00\%. This proves that our framework effectively suppresses the LLM's tendency to ask frequent, unnecessary questions, while successfully preserving its critical ability to recognize genuine information deficiency and seek help only when truly needed.

    \item \textbf{Superior Disambiguation (Intent \& Device Resolution):} For Intent Resolution (33.33\% $\rightarrow$ 70.83\%) and Device Resolution (46.15\% $\rightarrow$ 69.44\%), user instructions typically contain vague nouns requiring state-based deduction. SAGE's reactive ``think-while-acting'' loop often gets trapped in infinite API calls or hallucinates target parameters. In contrast, DS-IA's Stage 1 module acts as a semantic firewall, pre-aligning the vague instruction with the environment snapshot $S_t$ before generation, yielding massive absolute gains of +37.5\% and +23.29\% respectively.

    \item \textbf{Handling Complex Contexts (Personalization \& Persistence):} Personalization tasks involve subjective user preferences, where DS-IA exactly doubles the baseline success rate (27.27\% $\rightarrow$ 54.55\%) through stricter semantic routing. More impressively, on Persistence tasks that require long-term state monitoring, SAGE fundamentally struggles (25.00\%) due to context forgetfulness over multiple turns. DS-IA achieves a perfect 100.00\% Succ. Rate here, as its cascade verifier mechanically anchors every generated action to the physical truth, eradicating contextual hallucinations.

    \item \textbf{Foundational Capabilities (Simple \& Chained Commands):} Even on Simple Commands, SAGE often over-complicates the execution trace, leading to format collapse (57.14\%). DS-IA directly maps these to valid actions, increasing robustness to 85.71\%. Finally, for Command Chaining (purely sequential logic without deep ambiguity), both frameworks perform equally well (62.50\%), confirming that DS-IA introduces sophisticated disambiguation capabilities without degrading the baseline's foundational multi-step instruction-following competence.
\end{itemize}


\section{Conclusion}
\label{sec:conclusion}

This paper addresses two fundamental bottlenecks in deploying LLM-based agents for smart homes: the ``Reliability Gap'' (entity hallucinations) and the ``Interaction Frequency Dilemma'' (excessive user disturbance). We propose the Dual-Stage Intent-Aware (DS-IA) framework, which pioneers an ``Analyze-then-Act'' paradigm by decoupling macro-intent routing from micro-grounding validation.

\subsection{Summary of Contributions}
Our extensive evaluation on HomeBench and SAGE benchmarks yields three critical insights:
\begin{itemize}
    \item \textbf{Safety as a Priority:} DS-IA acts as an effective semantic firewall. By achieving an 87.04\% Exact Match (EM) on invalid instructions (HomeBench IS) compared to the baseline's 14.07\%, our framework demonstrates a robust capability for early rejection. This effectively eradicates the risk of ``Forced Grounding,'' ensuring physical safety even at the cost of slight recall loss on simple tasks.
    \item \textbf{Autonomy with Precision:} We successfully mitigate the interaction dilemma inherent in reactive frameworks. The proposed DS-IA framework improves the Autonomous Success Rate (the ability to resolve tasks independently without redundant user queries) from 42.86\% (SAGE) to 71.43\%. This significant enhancement demonstrates that State-Aware Disambiguation can effectively resolve most functional ambiguities by grounding intent in the physical environment, thereby minimizing unnecessary user disturbances while maintaining execution precision.
    \item \textbf{Long-Horizon Robustness:} The 100\% success rate on Persistence tasks demonstrates that our structured grounding mechanism is superior to reactive approaches in maintaining context over time.
\end{itemize}

\subsection{Limitations and Future Work}
Despite these advancements, we identify three directions for future optimization:
\begin{enumerate}
    \item \textbf{Multimodal Perception:} Currently, DS-IA relies on textual metadata. Inspired by \cite{enhancing2024} and \cite{iotlm}, future work will integrate Vision-Language Models (VLMs) to resolve visual references (e.g., \textit{``Turn on that red lamp''}) and process multi-sensory data.
    \item \textbf{Privacy-Preserving SLM Distillation:} To address inference latency and privacy concerns raised in \cite{privacy2025}, we plan to distill the Intent Analysis module into specialized Small Language Models (SLMs). This will enable on-device execution, eliminating the need to send private home states to the cloud.
    \item \textbf{Personalized Memory:} Following the findings in \cite{party2023}, we aim to incorporate a Vector Database (RAG) to memorize user habits, allowing the agent to implicitly resolve preference-based ambiguities.
\end{enumerate}

In conclusion, DS-IA offers a robust, safe, and efficient blueprint for the next generation of embodied IoT agents, bridging the gap between linguistic reasoning and physical execution.
\appendix

\section{Complete Prompt Templates for DS-IA Framework}
\label{app:prompt_templates}

To ensure full reproducibility and to demonstrate how physical grounding constraints are explicitly injected into the Large Language Models, Figure \ref{fig:full_prompt_templates} provides the exact system prompts and contextual structures utilized in both stages of our DS-IA framework.

\begin{figure*}[htbp]
\centering

\begin{tcolorbox}[
    colback=gray!5,           
    colframe=gray!50!black,   
    arc=4pt,                  
    boxrule=0.5pt,            
    left=8pt, right=8pt, top=8pt, bottom=8pt,
    fonttitle=\bfseries,
    title=(a) Prompt Template for Stage 1: Global Intent Analysis
]
\small
\textbf{\textsc{System Role:}} \\
You are a smart home intent analyzer. Your task is to strictly evaluate if the user's command can be executed based on the provided environment state.\\

\textbf{\textsc{Environment Context:}} \\
\texttt{<home\_state>} \\
\texttt{\{Device\_Status\_Snapshot\}} \\
\texttt{</home\_state>} \\

\textbf{\textsc{Evaluation Rules:}} \\
Check these THREE things for each operation:
\begin{enumerate}
    \setlength{\itemsep}{0pt}
    \setlength{\parskip}{0pt}
    \item \textbf{Room existence:} Does the mentioned room exist?
    \item \textbf{Device existence:} Does the device exist in that room?
    \item \textbf{Action support:} Does the device support the requested action/attribute?
\end{enumerate}

\textbf{\textsc{Output Format:}} \\
Output a JSON object containing an array of \texttt{operations} (with \texttt{valid} and \texttt{reason} fields) and a global \texttt{all\_valid} boolean flag.
\end{tcolorbox}

\vspace{2mm} 

\begin{tcolorbox}[
    colback=gray!5,           
    colframe=gray!50!black,   
    arc=4pt,                  
    boxrule=0.5pt,            
    left=8pt, right=8pt, top=8pt, bottom=8pt,
    fonttitle=\bfseries,
    title=(b) Prompt Template for Stage 2: Code Generation
]
\small
\textbf{\textsc{System Role \& Negative Constraints:}} \\
You are 'Al', a helpful AI Assistant that controls the devices in a house. Complete the following task as instructed or answer the following question with the information provided only. \\
The current status of the device and the methods it possesses are provided below, please only use the methods provided. \\
\textcolor{red!80!black}{\textbf{Output "error\_input" when operating non-existent attributes and devices. Only output machine instructions and enclose them in \{\}.}} \\

\textbf{\textsc{Environment Context \& Capabilities:}} \\
\texttt{<home\_state>} \\
The following provides the status of all devices in each room of the current household, the adjustable attributes... \\
\texttt{\{Device\_Status\_Snapshot\}} \\
\texttt{</home\_state>} \\

\texttt{<device\_method>} \\
The following provides the methods to control each device in the current household: \\
\texttt{\{Device\_Methods\_Snapshot\}} \\
\texttt{</device\_method>} \\

\textbf{\textsc{Few-Shot Demonstrations (Optional):}} \\
\texttt{\{In\_Context\_Examples\}} \\

\textbf{\textsc{User Interaction:}} \\
\texttt{-------------------------------} \\
\texttt{Here are the user instructions you need to reply to.} \\
\texttt{<User instructions:>} \\
\texttt{\{User\_Command\}} \\
\texttt{<Machine instructions:>}
\end{tcolorbox}

\caption{The complete constraint-aware prompt templates utilized in the DS-IA framework. (a) The Stage 1 prompt acts as a semantic firewall to extract diagnostic reasoning. (b) The Stage 2 prompt generates the final execution sequence, integrating strict negative constraints (\textit{e.g.,} utilizing \texttt{error\_input} for unexecutable actions) to enforce physical grounding.}
\label{fig:full_prompt_templates}
\end{figure*}

\end{document}